\documentclass[sigconf]{acmart}
\pdfoutput=1
\renewcommand\footnotetextcopyrightpermission[1]{} % removes footnote with conference information in first column
\usepackage{makecell}
\usepackage{multirow}
\usepackage{subfigure}
\usepackage{epstopdf}
\usepackage{amsmath,amsfonts}
\usepackage{stfloats}
\AtBeginDocument{%
  \providecommand\BibTeX{{%
    \normalfont B\kern-0.5em{\scshape i\kern-0.25em b}\kern-0.8em\TeX}}}

\begin{document}

\title{Generalizing Aggregation Functions in GNNs: High-Capacity GNNs via Nonlinear Neighborhood Aggregators}


\author{Beibei Wang and Bo Jiang\footnote{*}}
\affiliation{
	\institution{School of Computer Science and Technology of Anhui University}
	\city{Hefei}
	\country{China}
}
%\renewcommand{\shortauthors}{Anonymous Author, et al.}


\begin{abstract}
Graph neural networks (GNNs) have achieved great success in many graph learning tasks. The main aspect powering existing GNNs is the multi-layer network architecture to learn the nonlinear graph representations for the specific learning tasks.
The core operation in GNNs is message propagation in which each node updates its representation by aggregating its neighbors' representations.
Existing GNNs mainly adopt either {\emph{linear}} neighborhood  aggregation (mean, sum) or max aggregator in their message propagation.
(1) For linear aggregators, the whole nonlinearity and network's capacity of GNNs are generally limited due to deeper GNNs usually suffer from  over-smoothing issue.
%Most of existing GNNs
%adopt \emph{linear} neighborhood  aggregation mechanism and obtain \emph{nonlinear} representation via multiple activation functions in their layer-wise message propagation.
% Therefore,
%  the whole nonlinearity of GNNs is generally determined based on the number of hidden layers (depth of networks). % activation functions in hidden layers.
% However, it is known that, deeper GNNs usually suffer from  over-smoothing issue. Therefore, the nonlinearity and the whole learning capacity of linear aggregators based GNNs are generally limited.
(2) For max aggregator, it usually fails to be aware of the detailed information of node representations within neighborhood.
To overcome these issues,
we re-think the message propagation mechanism in GNNs and aim to develop the general nonlinear aggregators for neighborhood information aggregation in GNNs.
One main aspect of our proposed nonlinear aggregators is that they provide the optimally balanced aggregators between max and mean/sum aggregations. % which thus give optimally balanced aggregation strategies for GNNs.
Thus, our aggregators can inherit both (i) high nonlinearity that increases network's capacity and (ii) detail-sensitivity that preserves the detailed information of representations together in GNNs' message propagation.
%we re-think the message neighborhood aggregation mechanism in GNNs' layer-wise propagation and develop several \emph{nonlinear} neighborhood aggregation functions to enhance the network's capacity of GNNs.
%The proposed nonlinear aggregation mechanisms
% are general scheme which can be integrated with many GNN architectures to enhance existing GNNs's capacities.
% Moreover, our proposed nonlinear aggregation functions are intermediates between max and mean/summation aggregations which provide optimally balanced aggregation strategies for GNNs.
Promising experiments on several datasets show the effectiveness of the proposed nonlinear aggregators.
\end{abstract}

%\begin{CCSXML}
%<ccs2012>
%  <concept_id>10010520.10010553.10010562</concept_id>
%  <concept_desc>Computer systems organization~Embedded systems</concept_desc>
%  <concept_significance>500</concept_significance>
%  <concept_id>10010520.10010575.10010755</concept_id>
%  <concept_desc>Computer systems organization~Redundancy</concept_desc>
%  <concept_significance>300</concept_significance>
%  <concept_id>10010520.10010553.10010554</concept_id>
%  <concept_desc>Computer systems organization~Robotics</concept_desc>
%  <concept_significance>100</concept_significance>
%  <concept_id>10003033.10003083.10003095</concept_id>
%  <concept_desc>Networks~Network reliability</concept_desc>
%  <concept_significance>100</concept_significance>
%</ccs2012>
%\end{CCSXML}
%
%\ccsdesc[500]{Computer systems organization~Embedded systems}



\maketitle

\section{Introduction}
\label{submission}
\footnotetext[1]{*Corresponding author (Email: {jiangbo@ahu.edu.cn})}
% Compact learning on graph data is an active  problem in machine learning and data mining fields.
Graph Neural Networks (GNNs) have achieved great success in many graph data representation and learning tasks, such as semi-supervised learning, clustering, graph classification etc. As we know that the main aspect powering existing GNNs is the multi-layer network architecture to learn the rich nonlinear graph representations for the specific learning tasks.

It is known that the core message propagation mechanism in multi-layer GNNs is neighborhood information aggregation in which each node is updated by aggregating the information from its neighbors.
Most of existing GNNs generally adopt either {linear} neighborhood  aggregation  (\emph{e.g.}, mean, sum)~\cite{kipf2017semi,GIN} or max aggregation~\cite{hamilton2017inductive} or combination of them~\cite{Dehmamy2019,PNA} in their layer-wise message propagation. % and obtain the (nonlinear) representation via the activation function in each hidden layer.
For example,
Kipf et al.~\cite{kipf2017semi} present Graph Convolutional Networks (GCN) which adopts a weighted summation operation as the aggregation function.
Veli{\v c}kovi{\'c} et al.~\cite{velickovic2018graph} present Graph Attention Networks (GAT) which uses a learned weighted mean aggregator  via a self-attention mechanism.
Hamilton et al.~\cite{hamilton2017inductive} propose the general GraphSAGE  that employs mean, sum and max operation respectively for neighborhood aggregation.
Dehmamy et al.~\cite{Dehmamy2019} propose a modular GCN design by combining different neighborhood aggregation rules (mean, sum, normalized mean) together with residual connections.
Xu et al.~\cite{GIN} propose Graph Isomorphism Networks (GIN) in which the summation operation is utilized for neighborhood information aggregation for graph classification task.
Geisler et al.~\cite{GeislerZG20} propose reliable GNN by developing a robust aggregation function based on robust statistics in which the robust aggregator is  implemented via an adaptively weighted mean aggregation function.
Gabriele et al.~\cite{PNA} propose Principal Neighbourhood Aggregation (PNA) by integrating multiple aggregators (\emph{e.g.}, mean, max, min) together via  degree-scalers.
%Yang et al.~\cite{} propose Propagation-regularization (P-reg) to enhance the
%learning ability of existing GNNs by implementing message passing operation via mean aggregator.
Cai et al.~\cite{cai2021rethinking} propose Graph Neural Architecture Search (GNAS) to learn the optimal depth of message passing with max and sum neighbor aggregations.

%Cai et al.~\cite{cai2021rethinking}
%
%Zhu et al.~\cite{bgnn_ijcai20} propose Bilinear Graph Neural Networks (BGNN) by introducing element-wise product operation as the neighborhood aggregation.

%
%After reviewing the previous GNNs on various graph learning tasks, we can find that most of them adopt either \emph{linear aggregation function} or max aggregation as neighborhood information aggregator in layer-wise message propagation.

After reviewing the previous GNNs on various graph learning tasks, we can find the following aspects.
\emph{First}, most of existing GNNs generally adopt {linear aggregation functions} as neighborhood information aggregators in their layer-wise message propagation, i.e., they learn the context-aware representation of each node by linearly aggregating the information from its neighbors.
% These linear aggregator based GNNs mainly obtain nonlinear representations via the activation function $\sigma$ used in each hidden layer.
Thus, the whole nonlinearity of these \emph{linear aggregator} based GNNs is determined based on the number of hidden layers (depth of networks).
However, it is known that, deeper GNNs usually lead to over-smoothing issue~\cite{kipf2017semi,li2018deeper,PairNorm}. Therefore, the whole nonlinearity and network's capacity of these GNNs are generally limited.
\emph{Second}, the nonlinear max aggregator has been utilized in some works~\cite{hamilton2017inductive,lee2017generalizing,GIN}.
As we know, max aggregator obviously fails to preserve the detailed information of node representations within each node's neighborhood.
\emph{Third}, some recent works attempt to combine mean/sum and max aggregators together to provide a combined aggregator~\cite{PNA}.
However, the combined aggregator is explicitly depended on the individual mean/sum and max aggregators.

%
%Therefore, existing GNNs mainly obtain nonlinear representations via the activation function used in the hidden layers.
%Thus, the nonlinearity is determined based on the number of hidden layers (depth of networks). However, it is known that, deeper GNNs usually lead to over-smoothing issue~\cite{kipf2017semi,li2018deeper,PairNorm}. Therefore, the nonlinearity and the whole network's capacity of existing GNNs are still limited.

To address these issues, in this work, we re-think the message aggregation mechanism in layer-wise propagation of GNNs and aim to
fully exploit the flexible nonlinear aggregation functions for information aggregation in GNNs.
Specifically,
we develop three kinds of non-linear neighborhood aggregators for GNNs' message propagation by exploiting $L_p$-norm, polynomial and softmax functions respectively. %, for neighborhood information aggregation in GNNs.
Overall, there are three main aspects of the proposed methods.
\textbf{First},
one important property of the proposed nonlinear neighborhood aggregators is that they
%The main benefit of the proposed  non-learn message aggregation functions is that they enhance the learning capacity of GNNs. The proposed nonlinear aggregation functions
can be regarded as intermediates between linear mean/summation and nonlinear max aggregators and provide the optimal flexible \textbf{balanced} aggregation strategies for GNNs.
Thus, our aggregation mechanisms can  inherit both (i) \emph{nonlinearity} that increases network's functional complexity/capacity
and (ii) \emph{detail-sensitivity} that preserves the detailed information of representations in GNNs' message propagation.
\textbf{Second}, the proposed aggregators are all differentiable  that allow end-to-end training.
\textbf{Third}, our aggregation mechanisms are general scheme which can
integrate with  many GNN architectures to enhance existing GNNs' capacities and learning performance.
Overall, we summarize the main contributions as follows:
\begin{itemize}
	\item We propose to develop three kinds of nonlinear neighborhood aggregation schemes for the general GNNs' layer-wise propagation.
	\item We analyze the main properties of the proposed nonlinear neighborhood aggregators and show the balanced behavior of the proposed models.
	%     exploit three kinds of nonlinear functions a novel attentional mechanism, named Structural-aware Attention (SaATT), for structural graph data.
	\item We incorporate the proposed nonlinear aggregation mechanisms into GNNs and propose several novel networks for graph data representation and learning tasks.
\end{itemize}
We integrate our nonlinear aggregators into several GNN architectures and experiments on several datasets show the effectiveness of the proposed aggregators.

%------------------------------------------------------------------------
\section{Revisiting Neighborhood Aggregation in GNNs}

GNN provides a multi-layer network architecture for graph data representation and learning.
It is known that the core operation of GNNs is the neighborhood aggregation for network's layer-wise message propagation.
%Graph convolutional networks~\cite{kipf2017semi} is an important branch of GNNs for representation learning of graph structure data.
Let  $G(\mathbf{A}, \mathbf{H})$ be the input graph where
$\mathbf{A}\in \mathbb{R}^{n\times n}$ denotes the adjacency matrix and $\mathbf{H}=(\mathbf{H}_1, \mathbf{H}_2\cdots \mathbf{H}_n)\in \mathbb{R}^{n\times d}$ denotes the collection of node features. % and $\mathbf{A}\in \mathbb{R}^{n\times n}$ denotes the adjacency matrix.
% Then, % respectively, where
%$\mathbf{A}$ is the original adjacency matrix.
% $\mathbf{A}_{uv}=1$ denotes there exists an edge between the vertices $u$ and $v$,  and vice versa.
The neighborhood aggregation for layer-wise message propagation in GNNs~\cite{hamilton2017inductive,GeislerZG20} can generally be formulated as follows,
\begin{flalign}\label{EQ:Layer-wise}
	\mathbf{H}_{v}' = \sigma\Big(\mathbf{W}\cdot \mathrm{AGG}\big (\mathbf{A}_{vu},\mathbf{H}_{u},{u}\in \mathcal{N}_{v}\big )\Big)
\end{flalign}
where % $k=1, 2 \cdots K$ and $\mathbf{h}_{u}^{(0)}=\mathbf{h}_{u}, \forall{u}\in \mathcal{N}_{v}$ and
$\mathcal{N}_v$ denotes the neighborhood of node $v$ (including node $v$) and matrix $\mathbf{W}$ denotes the layer-wise weight parameter.
Function $\sigma(\cdot)$ denotes the activation function, such as $\mathrm{ReLU}$, $\mathrm{softmax}$, and $\mathrm{AGG}$ denotes the aggregation function, such as mean, summation, max, etc~\cite{hamilton2017inductive,GIN,GeislerZG20}.

% $\mathbf{h}_{u}^{(k)}, k=1, 2 \cdots K$ denotes the representation of node $u$ in the $k$-th layer and $\mathbf{h}_{u}^{(0)}=\mathbf{h}_{u}$. % is the input node feature.

For example,
in Graph Convolutional Networks (GCN)~\cite{kipf2017semi}, it adopts the weighted summation aggregation over normalized graph in its layer-wise propagation as,
\begin{flalign}\label{EQ:GCN}
	\mathrm{AGG}\big (\mathbf{A}_{vu},\mathbf{H}_{u},{u}\in \mathcal{N}_{v}\big ) = \sum_{u\in \mathcal{N}_v} \mathbf{\hat{A}}_{vu}\mathbf{H}_{u}
\end{flalign}
where $\hat{\mathbf{A}}=\tilde{\mathbf{D}}^{-\frac{1}{2}}\tilde{\mathbf{A}}\tilde{\mathbf{D}}^{-\frac{1}{2}}$ denotes the normalized adjacency matrix in which $\tilde{\mathbf{A}}=\mathbf{A}+\mathbf{I}$ and $\tilde{\mathbf{D}}$ is the diagonal matrix with $\tilde{\mathbf{D}}_{vv}=\sum_{u}\tilde{\mathbf{A}}_{vu}$.
In Graph Attention Network (GAT)~\cite{velickovic2018graph}, it first computes the attention $\alpha_{vu}$ for each graph edge as, %by defining the shared attentional mechanism $\varphi_{\theta}: \mathbb{R}^{d'\times d'}\rightarrow \mathbb{R}$ as, % attention $\alpha_{ij}$
\begin{align}\label{Eq:gat}
	\mathbf{\alpha}_{vu} = \mathrm{softmax}_{G}\big(\varphi_{\theta}(\textbf{H}_v \textbf{W}, \textbf{H}_u \textbf{W})\big)
	% & \textbf{h}'_i = \sigma \Big(\sum\nolimits_{j\in\mathcal{N}_i} \mathbf{\alpha}_{ij} \textbf{W}\textbf{h}_j\Big) %\\
	%
\end{align}
where $\textbf{W}\in \mathbb{R}^{d\times {d}'}$ denotes the  layer-specific  weight matrix and $\mathrm{softmax}_{G}$ denotes the softmax function defined on topological graph.
The function $\varphi_{\theta}(\cdot)$ denotes the learnable metric function parameterized by $\theta$.
The parameter $\theta$ is shared across different edges to achieve information communication.
%For example,
%in GATs~\cite{velickovic2018graph}, the function $\varphi_{\theta}(\cdot)$ is defined by using a single-layer feedforward neural network, parameterized by a weight vector $\theta$.
Based on the learned edge attention $\alpha_{vu}$, GAT then conducts the layer-wise information aggregation as, % define the node feature aggregation to obtain the context-aware feature representation  $\textbf{h}'_i$ for
% each node as, %nodes $\tilde{\textbf{H}}=(\tilde{\textbf{h}}_1,\tilde{\textbf{h}}_2\cdots \tilde{\textbf{h}}_n)$ as,
%
\begin{align}\label{Eq:gatupdate}
	\mathrm{AGG}\big (\mathbf{A}_{vu},\mathbf{H}_{u},{u}\in \mathcal{N}_{v}\big ) = \sum_{u\in\mathcal{N}_v} \mathbf{\alpha}_{vu} \textbf{H}_u
\end{align}
%
%where  $\textbf{W}\in \mathbb{R}^{d\times {d}'}$ denotes the network's layer-specific weight matrix. %  and $\sigma(\cdot)$ denotes the nonlinear activation function, such as $\mathrm{softmax}, \mathrm{ReLU}$, etc, to obtain nonlinear representations.
% and $\mathcal{N}_v$ denotes the neighborhood set of node $v$.
% Since the learned attention $\alpha$ is operated via softmax operation.
% Thus, GATs indeed adopts a mean aggregator
%
%
In GraphSAGE~\cite{hamilton2017inductive}, it employs the max-pooling operation over the neighborhood and implements layer-wise message aggregation as follows
\begin{equation}\label{EQ:GraphSAGE}
	\mathrm{AGG}\big (\mathbf{A}_{vu},\mathbf{H}_{u},{u}\in \mathcal{N}_{v}\big ) = \mathop\mathrm{max} \big (\big\{ \mathcal{F_\theta}(\mathbf{H}_{u}),u\in \mathcal{N}_v\big\} \big )
\end{equation}
where the $\mathcal{F_\theta}(\cdot)$ denotes the fully-connected network with an activation function and learnable parameter $\theta$. The function $\mathrm{max}$ denotes the element-wise max operation which aggregates the dimension-wise maximum value of node's neighbors.
%There are other common examples of $\mathbf{AGGREGATE}^{(k)}$ for GNNs, such as $\mathbf{MEAN}$\cite{chen2020iterative} and $\mathbf{MAX}$\cite{dai2018learning}.

\section{The Proposed Method}
Most of existing GNNs mainly obtain nonlinear representations via activation function $\sigma$, such as $\mathrm{ReLU}$, $\mathrm{softmax}$ etc, in their layer-wise message propagation.
Therefore, the nonlinearity of the whole network is determined based on the number of hidden layers.
However, it is known that, deeper GNNs usually lead to over-smoothing issue~\cite{kipf2017semi,li2018deeper,PairNorm}. Therefore, the nonlinearity and the whole learning capacity of existing GNNs are still limited.
Besides,  some GNNs use the nonlinear max aggregation as aggregator which fails to preserve the detailed information of node representations within each node's neighborhood.
To overcome these issues, we re-think the message aggregation mechanism in layer-wise propagation of GNNs and develop several nonlinear message aggregation functions to enhance the learning capacity of GNNs.
To be specific, we propose three kinds of nonlinear aggregations, i.e.,  $L_{p}$-norm, polynomial and softmax aggregation.
All these nonlinear aggregations provide balanced and self-adjust between mean and max functions, as discussed in detail below.
%In the following, we assume $G(\mathbf{A}, \mathbf{H})$ be the input graph where
%$\mathbf{A}\in \mathbb{R}^{n\times n}$ denotes the adjacency matrix and $\mathbf{H}=(\mathbf{h}_1, \mathbf{h}_2\cdots \mathbf{h}_n)\in \mathbb{R}^{n\times d}$ denotes the collection of node features.

\subsection{$L_{p}$-norm Aggregation}
As a general nonlinear function, $L_{p}$-norm has been commonly used in computer vision and signal processing fields~\cite{Learned-NormPooling,lp-pooling}.
%$L_{p}$-norm function
Here, we propose to exploit $L_{p}$-norm function for the neighborhood aggregation in GNN's layer-wise message propagation.

Given $G(\mathbf{A}, \mathbf{H})$ be the input graph where
$\mathbf{A}\in \mathbb{R}^{n\times n}$ denotes the adjacency matrix and $\mathbf{H}=(\mathbf{H}_1, \mathbf{H}_2\cdots \mathbf{H}_n)\in \mathbb{R}^{n\times d}$ denotes the collection of node features.
One  straightforward  way to adopt $L_p$-norm for the neighborhood aggregation can be formulated as
\begin{equation}\label{EQ:GNN-LP}
	\mathrm{AGG}_{\ell_p}\big (\mathbf{A}_{vu},\mathbf{H}_{u},{u}\in \mathcal{N}_{v}\big ) = \Big(\sum_{u\in \mathcal{N}_v} \mathbf{{A}}_{vu}\big|\mathbf{H}_u\big|^p\Big)^{1/p}
\end{equation}
Here, $\big|\mathbf{H}_u\big|^p\in \mathbb{R}^d$ is defined as
$$
\big|\mathbf{H}_u\big|^p = \left(\big|\mathbf{H}_{u1}\big|^p, \big|\mathbf{H}_{u1}\big|^p\cdots \big|\mathbf{H}_{u1}\big|^p\right)
$$
where %$$ and % $\mathcal{N}_{v}$ denotes the neighborhood of node $v$.
$p\in [1, +\infty)$ is a parameter which can be learned adaptively.
However, the negative information has been ignored in Eq.(\ref{EQ:GNN-LP}).
To address this issue,
we adapt the above $L_p$-norm function and propose our aggregation  as follows,
%
%The general formulation of Formally, our $L_{p}$-norm based aggregation is formulated as follows,
% Graph Convolutional Networks (GCN)~\cite{kipf2016semi} adopts the sum aggregation in its layer-wise propagation as,
%\begin{equation}\label{EQ:GNN-LP}
%\mathrm{AGG}_{\ell_p}\big (\mathbf{A}_{vu},\mathbf{h}_{u},{u}\in \mathcal{N}_{v}\big ) = \Big(\sum_{u\in \mathcal{N}_v} \mathbf{\hat{A}}_{vu}\mathbf{h}^p_{u}\Big)^{1/p}
%\end{equation}
%where $p\in [0, +\infty)$ is a parameter and denotes the order of the norm.
%Note that, for the inevitable negative input value, we then reformulate Eq.(\ref{EQ:GNN-LP}) as below,
\begin{equation}\label{EQ:GNN-LP1}
	\mathrm{AGG}_{\ell_p} =
	\Big(\sum_{u\in \mathcal{N}_v} \mathbf{{A}}_{vu}{\big(\mathbf{H}_{u}-\mu_{m}\big)}^p\Big)^{1/p} + {\mu}_{m}
\end{equation}
where $\mu_m = \min_{u,k}\big\{\textbf{H}_{uk}\big\}$ denotes the minimum element  of matrix $\textbf{H}$ and $\textbf{H}_{u}\in \mathbb{R}^d$ denotes the feature vector of node $u$.  

\textbf{Remark. }
The proposed $\mathrm{AGG}_{\ell_p}$ is a general scheme which can be integrated with many GNN' architectures, such as GCN~\cite{kipf2017semi}, GAT~\cite{velickovic2018graph}, GraphSAGE~\cite{hamilton2017inductive}, etc.
For example, one can replace  $\mathbf{{A}}_{vu}$ with ${\mathbf{\hat{A}}}_{vu}$ in Eq.(2) to produce a GCN variant~\cite{kipf2017semi}.
Also, one can use the learned graph attention $\alpha_{vu}$ in Eq.(4) to replace $\mathbf{{A}}_{vu}$  to generate a new GAT model.
In particular, one main property of the proposed $\mathrm{AGG}_{\ell_p}$ is that it provides a balanced aggregator between max aggregator and  summation aggregator. Specifically, we have the following  Proposition 1. \\

\textbf{Proposition 1}:
\emph{When $p=1$, $\mathrm{AGG}_{\ell_p}$ becomes to the weighted summation aggregation; When $p \to \infty$, it is equivalent to the max aggregation.}\\

\textbf{Proof.}
When $p=1$, one can easily observe that $\mathrm{AGG}_{\ell_p}$ becomes to the weighted summation aggregation.
For the case $p \to \infty$, we first prove the following conclusion. % of the vector form.

To be specific, given any vector $x$ and $a\in \mathbb{R}^n, a_i\geq 0$, without loss of generality, we assume $x_m$ be the unique maximum value of $x$. % where $m$ is the indictor.
Below, we can show that
\begin{align}
	&\lim_{p \to \infty}\Big(\sum^n_{i=1} a_i|x_i|^p\Big)^{1/p}=\lim_{p\to \infty}\Big(|x_m|^p\sum^n_{i=1}a_i\big(\frac{|x_i|}{|x_m|}\big)^p\Big)^{1/p}\nonumber \\
	&=|x_m|\lim_{p \to \infty}\Big(\sum^n_{i=1} a_i\big(\frac{|x_i|}{|x_m|}\big)^p\Big)^{1/p}
	%&=|x_m|\lim_{p \to \infty}\big( a_m\big)^{1/p}=|x_m|
\end{align}
%When $p \to \infty$, $\mathrm{AGG}_{\ell_p}$
For any $i\neq m$, since $\frac{|x_i|}{|x_m|}<1$, thus $\lim_{p \to \infty}\big(\frac{|x_i|}{|x_m|}\big)^p=0$.
Thus, we have
\begin{equation}
	%&\lim_{p \to \infty}\Big(\sum^n_{i=1} a_i|x_i|^p\Big)^{1/p}=\lim_{p\to \infty}\Big(|x_m|^p\sum^n_{i=1}a_i\big(\frac{|x_i|}{|x_m|}\big)^p\Big)^{1/p}\nonumber \\
	|x_m|\lim_{p \to \infty}\Big(\sum^n_{i=1} a_i\big(\frac{|x_i|}{|x_m|}\big)^p\Big)^{1/p}=|x_m|\lim_{p \to \infty}\big( a_m\big)^{1/p}=|x_m|
\end{equation}
Based on the above analysis,
one can see that
\begin{equation}
	\lim_{p\to \infty}\Big(\sum_{u\in \mathcal{N}_v} \mathbf{{A}}_{vu}{(\mathbf{H}_{u}-\mu_{m})}^p\Big)^{1/p} + {\mu}_{m} = \max_{k\in\{1,2\cdots d\}}\big\{\textbf{H}_{uk} \big\}
\end{equation}
This completes the proof.

\subsection{Polynomial Aggregation}
In addition to $L_p$-norm, we also explore polynomial function~\cite{p-pooling} for neighborhood information aggregation.

Given $G(\mathbf{A}, \mathbf{H})$ be the input graph where
$\mathbf{A}\in \mathbb{R}^{n\times n}$ denotes the adjacency matrix and $\mathbf{H}=(\mathbf{H}_1, \mathbf{H}_2\cdots \mathbf{H}_n)\in \mathbb{R}^{n\times d}$ denotes the node features.
Formally, we propose to define polynomial aggregation $\mathrm{AGG}_{\mathcal{P}}$ as
\begin{equation}\label{EQ:GNN-P}
	\mathrm{AGG}_{\mathcal{P}}\big (\mathbf{A}_{vu},\mathbf{H}_{u},{u}\in \mathcal{N}_{v}\big ) =
	\dfrac{\sum\limits_{u\in \mathcal{N}_v} \mathbf{{A}}_{vu}|\mathbf{H}_{u}|^{\alpha+1}}{\sum\limits_{u\in \mathcal{N}_v} \mathbf{{A}}_{vu}|\mathbf{H}_u|^{\alpha}}
\end{equation}
where the division operation in Eq.(11) is the element-wise division and $|\mathbf{H}_u|^{\alpha}\in \mathbb{R}^d$ is defined as
$$
\big|\mathbf{H}_u\big|^{\alpha} = \big(\big|\mathbf{H}_{u1}\big|^{\alpha}, {\big|\mathbf{H}_{u1}\big|}^{\alpha}\cdots \big|\mathbf{H}_{u1}\big|^{\alpha}\big)
$$
where $\alpha\in [0,+\infty)$ is a parameter which controls its polynomial order. In this paper, the parameter value of $\alpha$ is learned adaptively. Similar to the above $L_p$-norm function, considering the inevitable negative elements in $\textbf{H}$ and the power operation~\cite{p-pooling}, we adapt the Eq.(\ref{EQ:GNN-P}) and finally propose our $\mathrm{AGG}_{\mathcal{P}}$ as follows,
\begin{equation}\label{EQ:GNN-P1}
	\mathrm{AGG}_{\mathcal{P}} =
	\frac{\sum\limits_{u\in \mathcal{N}_v} \mathbf{{A}}_{vu}{\big(\mathbf{H}_{u}-\mu_m\big)}^{\alpha+1}}{\sum\limits_{u\in \mathcal{N}_v} \mathbf{{A}}_{vu}{\big(\mathbf{H}_{u}-\mu_{m}\big)}^\alpha}+\mu_{m}
\end{equation}
where $\mu_m$ is the minimum element of $\textbf{H}$ and the division operation is the element-wise division.
%\textbf{Remark. }

Similar to $\mathrm{AGG}_{\ell_p}$, $\mathrm{AGG}_{\mathcal{P}}$ also gives  a general scheme and can be integrated with many GNNs' architectures, such as GCN~\cite{kipf2017semi}, GAT~\cite{velickovic2018graph}, GraphSAGE~\cite{hamilton2017inductive}, etc.
%To be specific, we can use $\hat{\mathbf{{A}}}_{vu}$ in Eq.(2) to replace  $\mathbf{{A}}_{vu}$  to produce a GCN variant~\cite{kipf2017semi}.
%We can also use the learned graph attention $\alpha_{vu}$ to replace $\mathbf{{A}}_{vu}$  in Eq.(4) to generate a new GAT model.
In addition, $\mathrm{AGG}_{\mathcal{P}}$ also provides a balanced aggregator between max and  summation aggregator~\cite{p-pooling}.
Formally, we have the following Proposition 2. \\

\textbf{Proposition 2}:
\emph{When $\alpha=0$, $\mathrm{AGG}_{\mathcal{P}}$ becomes to the weighted mean aggregation; When $\alpha \to +\infty$, it is equivalent to the max aggregation.}\\

The mathematical proof of this Proposition 2 can be similarly obtained by referring to work~\cite{p-pooling}. % for the mathematical proof of this proposition. % can be  presented in the work~\cite{}. %supplementary file.

\subsection{Softmax Aggregation}

The nonlinear function softmax has been commonly used as activity function in GNNs.
Given a vector $x\in \mathbb{R}^n$, the standard softmax function is defined as follows,
\begin{equation}\label{EQ:Softmax}
	\mathrm{Softmax}(x)_{i}=  \frac{{e}^{\gamma x_i}}{\sum\limits_{i=1}^n {e}^{\gamma x_i}} , i=1,2\cdots n %\,\,\,\,\, where \,\,\,\, x = (x_1, \cdots, x_n).
\end{equation}
where $\gamma >0$ is a scale factor.
It is known that softmax is a smooth approximation of argmax function, i.e., let $x_m$ be the maximum value of $x$, then
\begin{align}\label{EQ:Softmax}
	\lim_{\gamma \to +\infty} \mathrm{Softmax}(x)_i =  \left\{\begin{aligned} &1, \,\,\, i=m \\ &0,\,\,\,otherwise \end{aligned}\right.% \mathrm{Softmax}(x)_{i}=  \frac{{e}^{\gamma x_i}}{\sum\limits_{i=1}^n {e}^{\gamma x_i}}% , \,\,\,\,\, where \,\,\,\, x = (x_1, \cdots, x_n).
\end{align}
%
%One can extend the standard softmax function by a scale factor $\gamma$ and be formulated as
%\begin{equation}\label{EQ:Softmax1}
%\mathrm{Softmax(x)_{i}}=  \frac{{e}^{\gamma x_i}}{\sum\limits_{i=1}^n {e}^{\gamma x_i}}, \,\,\,\,\, where \,\,\,\, x = (x_1, \cdots, x_n).
%\end{equation}

In our work, we exploit softmax function for neighborhood information aggregation in GNNs.
To be specific, given $G(\mathbf{A}, \mathbf{H})$ be the input graph where
$\mathbf{A}\in \mathbb{R}^{n\times n}$ denotes the adjacency matrix and $\mathbf{H}=(\mathbf{H}_1, \mathbf{H}_2\cdots \mathbf{H}_n)\in \mathbb{R}^{n\times d}$ denotes the node features.
Formally, our softmax based aggregator $\mathrm{AGG}_{\mathcal{S}}$ is defined as
\begin{equation}\label{EQ:GNN-Softmax}
	\mathrm{AGG}_{\mathcal{S}}= \sum_{u\in \mathcal{N}_v} \mathbf{{A}}_{vu}\Big(\textbf{H}_u\odot \frac{{e}^{\gamma  \mathbf{H}_{u}}}{\sum\limits_{u\in \mathcal{N}_v} {e}^{\gamma  \mathbf{H}_{u}}}\Big)
\end{equation}
where $\gamma\in [0,+\infty)$ denotes the scalar parameter and
$\odot$ indicates the element-wise product. $\mathbf{H}_{u}\in \mathbb{R}^d$ denotes the
feature vector of node $u$.
The division operation in Eq.(15) denotes the element-wise division.
The operation $e^{\gamma  \mathbf{H}_{u}}$ is defined as
$$
e^{\gamma  \mathbf{H}_{u}} = \big( e^{\gamma  \mathbf{H}_{u1}}, e^{\gamma  \mathbf{H}_{u2}}\cdots e^{\gamma  \mathbf{H}_{ud}} \big)
$$
%
% \textbf{Remark. }
% $\mathrm{AGG}_{\mathcal{S}}$ can be regarded as  a general message aggregation scheme with a weight factor.
 Similar to the above two nonlinear  aggregators, $\mathrm{AGG}_{\mathcal{S}}$ can also combine with some GNNs' architectures, such as GCN~\cite{kipf2017semi} and GAT~\cite{velickovic2018graph}, etc.
In addition, we can show that $\mathrm{AGG}_{\mathcal{S}}$ provides an adaptive aggregator between max and mean aggregations.
Formally, we have the following proposition. \\

\textbf{Proposition 3}:
\emph{When $\gamma=0$, $\mathrm{AGG}_{\mathcal{S}}$ becomes to the weighted mean aggregation; %When $p=1$, $\mathrm{AGG}_{\mathcal{S}}$ becomes to softmax aggregation;
	When $\gamma \to \infty$, it is equivalent to the max aggregation.}\\

When $\gamma=0$, $\mathrm{AGG}_{\mathcal{S}}$ obviously becomes to the weighted mean aggregator.
For $\gamma \to \infty$, the proof can be easily obtained by using Eq.(14).

\begin{table*}[!htpb]
%	\Large
	\centering
	\caption{The introduction and usage of experimental datasets.}
	\centering
	\begin{tabular}{c c c c c c c}
		\hline
		\hline
		Dataset & Cora & Citeseer & Pubmed  & Amazon Photo & Amazon Computers & PPI \\
		\hline
		Nodes  & 2708 & 3327 & 19717 & 7487 &13381 &56944\\
		%\hline		
		Feature   & 1433 & 3703 & 500 & 745 &767 & 50 \\		
		%\hline
		Edges    & 5429 & 4732 & 44338 & 119043 &245778 &266144 \\
		%\hline
		Classes & 7 & 6 & 3 & 8 & 10& 121\\
		\hline
		\hline
	\end{tabular}
\end{table*}
%%%
\begin{table*}[!ht]
%	\Large
	\centering
	\caption{\upshape Results of semi-supervised classification for transductive learning task.
	}
	\begin{tabular}{ c c c c c c c}
		\hline
		\hline		
		&Method & Cora & Citeseer & Pubmed &Photo &Computers\\
		\hline
		&GCN (baseline) & 81.24$\pm$1.68  &69.92$\pm$1.75 & 79.82$\pm$1.60 & 89.05$\pm$1.80 & 80.02$\pm$2.83\\
		&GCN-$\mathrm{AGG}_{\ell_p}$ &82.57$\pm$1.88  &71.03$\pm$0.74 &80.72$\pm$1.00 &90.86$\pm$1.06 & 81.98$\pm$2.27\\
		&GCN-$\mathrm{AGG}_{\mathcal{P}}$ &82.20$\pm$1.60  &71.12$\pm$1.10 &81.01$\pm$1.29 & 91.80$\pm$0.50 & 82.86$\pm$1.41\\
		&GCN-$\mathrm{AGG}_{\mathcal{S}}$ &83.10$\pm$1.28  &71.50$\pm$1.12 &81.50$\pm$0.82 &91.25$\pm$0.80 &82.02$\pm$2.11 \\	
		\hline
		&Masked GCN (baseline)& 81.44$\pm$1.19  &69.10$\pm$1.98 & 80.19$\pm$1.40 & 88.60$\pm$3.08 & 76.06$\pm$7.09\\
		&Masked GCN-$\mathrm{AGG}_{\ell_p}$ &82.56$\pm$1.45  &69.96$\pm$1.33 &80.92$\pm$1.18 &90.56$\pm$1.19 & 80.63$\pm$4.92\\
		&Masked GCN-$\mathrm{AGG}_{\mathcal{P}}$ &82.00$\pm$1.02  &69.94$\pm$1.23 &81.03$\pm$0.81 & 90.98$\pm$1.95 & 78.15$\pm$6.44\\
		&Masked GCN-$\mathrm{AGG}_{\mathcal{S}}$ &82.88$\pm$1.00  &70.10$\pm$1.04 &81.50$\pm$1.06 &90.57$\pm$1.56 &81.50$\pm$2.75 \\	
		\hline
		&GAT (baseline) &82.02$\pm$1.55  &70.88$\pm$1.36 &80.05$\pm$0.97 & 89.72$\pm$2.02 & 79.72$\pm$2.20 \\
		&GAT-$\mathrm{AGG}_{\ell_p}$ &83.22$\pm$0.78 & 72.24$\pm$0.99 & 80.49$\pm$1.14 &  90.37$\pm$2.10 & 81.14$\pm$1.80\\
		&GAT-$\mathrm{AGG}_{\mathcal{P}}$ &83.30$\pm$1.15 & 71.56$\pm$1.46 & 80.29$\pm$1.05  & 90.70$\pm$1.49 &82.60$\pm$1.62 \\
		&GAT-$\mathrm{AGG}_{\mathcal{S}}$ &83.14$\pm$1.30 &72.15$\pm$1.01 & 81.05$\pm$1.22 & 90.65$\pm$2.10 & 81.26$\pm$2.10\\
		\hline
		&CAT (baseline) &82.45$\pm$1.40  &72.46$\pm$1.33 &80.12$\pm$1.10 & 88.04$\pm$2.14 & 80.53$\pm$1.85\\
		&CAT-$\mathrm{AGG}_{\ell_p}$&82.93$\pm$1.30  &72.50$\pm$0.87 &81.52$\pm$1.28 & 90.54$\pm$1.08 & 81.10$\pm$1.92\\
		&CAT-$\mathrm{AGG}_{\mathcal{P}}$&82.60$\pm$1.45  &71.78$\pm$1.15 &81.35$\pm$1.91 & 90.65$\pm$0.81 &82.66$\pm$1.43 \\
		&CAT-$\mathrm{AGG}_{\mathcal{S}}$&83.34$\pm$1.41  &72.61$\pm$0.80 & 82.04$\pm$1.40&91.21$\pm$0.66  &82.90$\pm$1.98 \\
		\hline
		\hline
	\end{tabular}
\end{table*}
%-------------------------------------------------------
\section{Experiments}
To validate the effectiveness of our proposed nonlinear aggregators, we take two GCN-based models (GCN~\cite{kipf2017semi} and Masked GCN~\cite{ijcai2019-565}) and  two GAT-based models (GAT~\cite{velickovic2018graph} and CAT~\cite{cat}) as baseline  architectures and perform evaluations on several widely used graph learning datasets. %including semi-supervised node classification and unsupervised node clustering.

\subsection{Experimental Settings}
\noindent\textbf{Dataset Setting.}
We test the proposed models on six datasets including Cora, Citeseer, Pubmed~\cite{sen2008collective}, PPI~\cite{hamilton2017inductive}, Amazon Photo and Computers~\cite{shchur2018pitfalls}.
%For citation datasets, nodes represent the documents and edges indicate the connection relationships among documents. Each node is described by a bag of word vector.
Similar to work~\cite{kipf2017semi}, we randomly select $20$ nodes per class as training set, $500$ nodes and $1000$ nodes as validation and testing set respectively.
%For amazon datasets, nodes denote the goods and edges encode that goods are often purchased together. Each node is described by a bag of word vector.
For Amazon Photo and Computers~\cite{shchur2018pitfalls}, following work~\cite{shchur2018pitfalls}, we randomly select $20$ nodes per class as training set, $30$ nodes per class and the remaining nodes as validation and testing set respectively.
For PPI dataset, % each node indicates a protein and edge indicates whether two proteins interact with each other. It contains $24$ graph corresponding to different human tissues.
similar to the setting in work~\cite{velickovic2018graph}, we take $20$ graphs as training set, $2$ graphs as validation set and $2$ graphs as testing set.
The introduction and usage of these datasets are summarized in Table 1.

\noindent\textbf{Parameter Setting.}
We integrate the proposed nonlinear aggregators with GCNs and GATs respectively.
For GCN-based methods,
the number of hidden layer units is selected from ${16, 64}$ for all datasets.
For GAT-based methods,
the number of hidden layer units is set to $8$ on all datasets.
% The rest parameters such as learning rate, dropout rate refer to the setting of GAT~\cite{velickovic2018graph}.
The weight decay are set to $5e-4$ and $1e-5$ for citation and amazon datasets, respectively.
Our network parameters $\{W, p/\alpha/\gamma\}$ are trained and optimized by gradient descent algorithms~\cite{Adam,momentum}. % We then select the optimal model for testing according to the validation set.
%

%%%
\begin{table*}[!ht]
%	\Large
	\centering
	\caption{\upshape Results of semi-supervised clustering on five datasets.
	}
	\begin{tabular}{ c c c c c c c}
		\hline
		\hline		
		&Method & Cora & Citeseer & Pubmed &Photo &Computers\\
		\hline
		&GCN (baseline)& 80.24$\pm$0.97  &70.15$\pm$0.60 & 79.10$\pm$1.22 & 89.00$\pm$1.45 & 80.36$\pm$2.40\\
		&GCN-$\mathrm{AGG}_{\ell_p}$ &82.40$\pm$0.82  &70.58$\pm$0.83 &79.98$\pm$1.21 &91.04$\pm$1.10 & 82.43$\pm$1.95\\
		&GCN-$\mathrm{AGG}_{\mathcal{P}}$ &82.14$\pm$0.95 &71.00$\pm$0.89 &79.96$\pm$0.98 & 92.06$\pm$0.33 & 83.14$\pm$1.38\\
		&GCN-$\mathrm{AGG}_{\mathcal{S}}$ &82.87$\pm$0.91  &71.28$\pm$0.82 &80.64$\pm$0.89 &91.22$\pm$0.82 &82.53$\pm$1.82 \\	
		\hline
		&Masked GCN (baseline)& 80.74$\pm$0.78  &69.03$\pm$1.78 & 79.14$\pm$1.05 & 88.83$\pm$3.03 & 76.19$\pm$7.25\\
		&Masked GCN-$\mathrm{AGG}_{\ell_p}$ &82.40$\pm$0.66  &69.76$\pm$1.07 &80.26$\pm$0.85 &90.76$\pm$1.16 & 80.92$\pm$5.00\\
		&Masked GCN-$\mathrm{AGG}_{\mathcal{P}}$ &81.90$\pm$0.42  &69.83$\pm$1.02 &80.00$\pm$1.06 & 91.15$\pm$1.90 & 78.74$\pm$6.28\\
		&Masked GCN-$\mathrm{AGG}_{\mathcal{S}}$ &82.58$\pm$0.85  &69.96$\pm$1.08 &80.76$\pm$1.26 &90.79$\pm$1.52 &81.87$\pm$2.69 \\	
		\hline
		&GAT (baseline) &82.55$\pm$0.40  &71.02$\pm$0.30 &80.01$\pm$0.71 & 88.55$\pm$0.70 & 78.98$\pm$1.20 \\
		&GAT-$\mathrm{AGG}_{\ell_p}$ &83.62$\pm$0.56 & 72.30$\pm$0.45 & 80.40$\pm$1.16 &  90.61$\pm$1.33 & 81.93$\pm$1.91\\
		&GAT-$\mathrm{AGG}_{\mathcal{P}}$ &83.68$\pm$0.65 & 71.76$\pm$1.10 & 80.22$\pm$0.80  & 90.78$\pm$1.18 &81.06$\pm$1.20 \\
		&GAT-$\mathrm{AGG}_{\mathcal{S}}$ &83.50$\pm$0.65 &71.73$\pm$0.76 & 80.89$\pm$1.10 & 90.84$\pm$1.15 & 80.54$\pm$1.94\\
		\hline
		&CAT (baseline) &81.98$\pm$1.01  &70.35$\pm$0.72 &77.85$\pm$3.48 & 89.67$\pm$1.16 & 80.55$\pm$1.69\\
		&CAT-$\mathrm{AGG}_{\ell_p}$&82.59$\pm$0.82  &71.68$\pm$0.66 &79.75$\pm$1.27 & 90.72$\pm$1.05 & 81.88$\pm$1.78\\
		&CAT-$\mathrm{AGG}_{\mathcal{P}}$&82.10$\pm$1.10  &70.49$\pm$0.85 &79.92$\pm$1.16 &90.83$\pm$0.77  &83.00$\pm$1.39 \\
		&CAT-$\mathrm{AGG}_{\mathcal{S}}$&82.60$\pm$1.03 &71.78$\pm$0.75 &80.40$\pm$1.24 &91.42$\pm$0.61  &83.25$\pm$1.94 \\
		\hline
		\hline
	\end{tabular}
\end{table*}
%%%---------------------------

\subsection{Comparison Results}

\noindent\textbf{Node Classification.}
% We evaluate the proposed methods on semi-supervised node classification task.
We first test our proposed models on transductive learning task and take some popular GNNs as baselines including Graph Convolutional Network (GCN)~\cite{kipf2017semi},  Graph Attention Networks (GAT)~\cite{velickovic2018graph}, Masked GCN~\cite{ijcai2019-565} and CAT~\cite{cat}.
Table 2 reports the comparison results. One can observe that the proposed nonlinear aggregators can consistently improve the baseline models which demonstrates the effectiveness of our proposed nonlinear aggregation schemes on extending the network's capacity and thus  enhancing GNNs' learning performance.
We then test our proposed models on inductive learning task. % and compare with some recent related GNNs including GCN~\cite{kipf2017semi}, GAT ~\cite{velickovic2018graph}, CAT~\cite{cat} and Masked GCN~\cite{ijcai2019-565}.
 Table 4 reports the comparison results. We can note that the nonlinear GNN models achieve better result than vanilla GNNs which further indicates the advantages of our proposed nonlinear aggregators.
\begin{table}[!htpb]
%	\Large
	\centering
	\caption{Results of semi-supervised classification for inductive learning task.}
	\centering
	\begin{tabular}{c c}
		\hline
		\hline
		Method    & PPI\\
		\hline
		GCN (baseline)  &97.25$\pm$0.34 \\
		GCN-$\mathrm{AGG}_{\ell_p}$ &98.64$\pm$0.01\\
		GCN-$\mathrm{AGG}_{\mathcal{P}}$ &98.60$\pm$0.02 \\
		GCN-$\mathrm{AGG}_{\mathcal{S}}$ &98.70$\pm$0.44\\
		\hline
		Masked GCN (baseline) &97.30$\pm$0.50 \\
		Masked GCN-$\mathrm{AGG}_{\ell_p}$ & 98.41$\pm$0.18 \\
		Masked GCN-$\mathrm{AGG}_{\mathcal{P}}$ &98.52$\pm$0.11 \\
		Masked GCN-$\mathrm{AGG}_{\mathcal{S}}$ & 97.95$\pm$0.29 \\
		\hline
		GAT (baseline) &97.30$\pm$0.30 \\
		GAT-$\mathrm{AGG}_{\ell_p}$ &98.22$\pm$0.60\\
		GAT-$\mathrm{AGG}_{\mathcal{P}}$ &98.60$\pm$0.40 \\
		GAT-$\mathrm{AGG}_{\mathcal{S}}$ &98.54$\pm$0.70 \\
		\hline
		CAT (baseline)  &97.80$\pm$0.25 \\
		CAT-$\mathrm{AGG}_{\ell_p}$ & 98.32$\pm$0.13 \\
		CAT-$\mathrm{AGG}_{\mathcal{P}}$ & 98.44$\pm$0.08 \\
		CAT-$\mathrm{AGG}_{\mathcal{S}}$ & 98.12$\pm$0.17 \\
		\hline
		\hline
	\end{tabular}
\end{table}
%%-----------------------------------------------------------

\noindent\textbf{Node Clustering.}
We also evaluate the proposed methods on semi-supervised node clustering task as suggested in work~\cite{cat}. % and take some popular GNNs as baselines including Graph Convolutional Network (GCN)~\cite{kipf2017semi},  Graph Attention Networks (GAT)~\cite{velickovic2018graph}, Masked GCN~\cite{ijcai2019-565} and CAT~\cite{cat}.
Table 3 reports the comparison results. Note that the nonlinear aggregation based  models obtain higher performance which shows the benefits of our proposed nonlinear message aggregation mechanisms for GNNs' learning.

%%%%
%\begin{figure*}[!ht]
%	\centering
%	\subfigure
%	{   \begin{minipage}{5.7cm}
%			\centering
%			\includegraphics[width=1.\textwidth]{cora_depth.eps}
%		\end{minipage}
%	}
%	\subfigure
%	{   \begin{minipage}{5.7cm}
%			\centering
%			\includegraphics[width=1.\textwidth]{citeseer_depth.eps}
%		\end{minipage}
%	}
%	\subfigure
%	{   \begin{minipage}{5.7cm}
%			\centering
%			\includegraphics[width=1.\textwidth]{pubmed_depth.eps}
%		\end{minipage}
%	}												
%	\caption{Different hidden layers of GCN~\cite{kipf2017semi}, GCN-$\mathrm{AGG}_{\ell_p}$, GCN-$\mathrm{AGG}_{\mathcal{P}}$ and GCN-$\mathrm{AGG}_{\mathcal{S}}$ on citation datasets.}\label{fig::depth}
%\end{figure*}
%%%%%
%%%%
\begin{figure*}[!ht]
	\centering
	\subfigure[GCN]
	{   \begin{minipage}{4.1cm}
			\centering
			\includegraphics[width=1.0\textwidth]{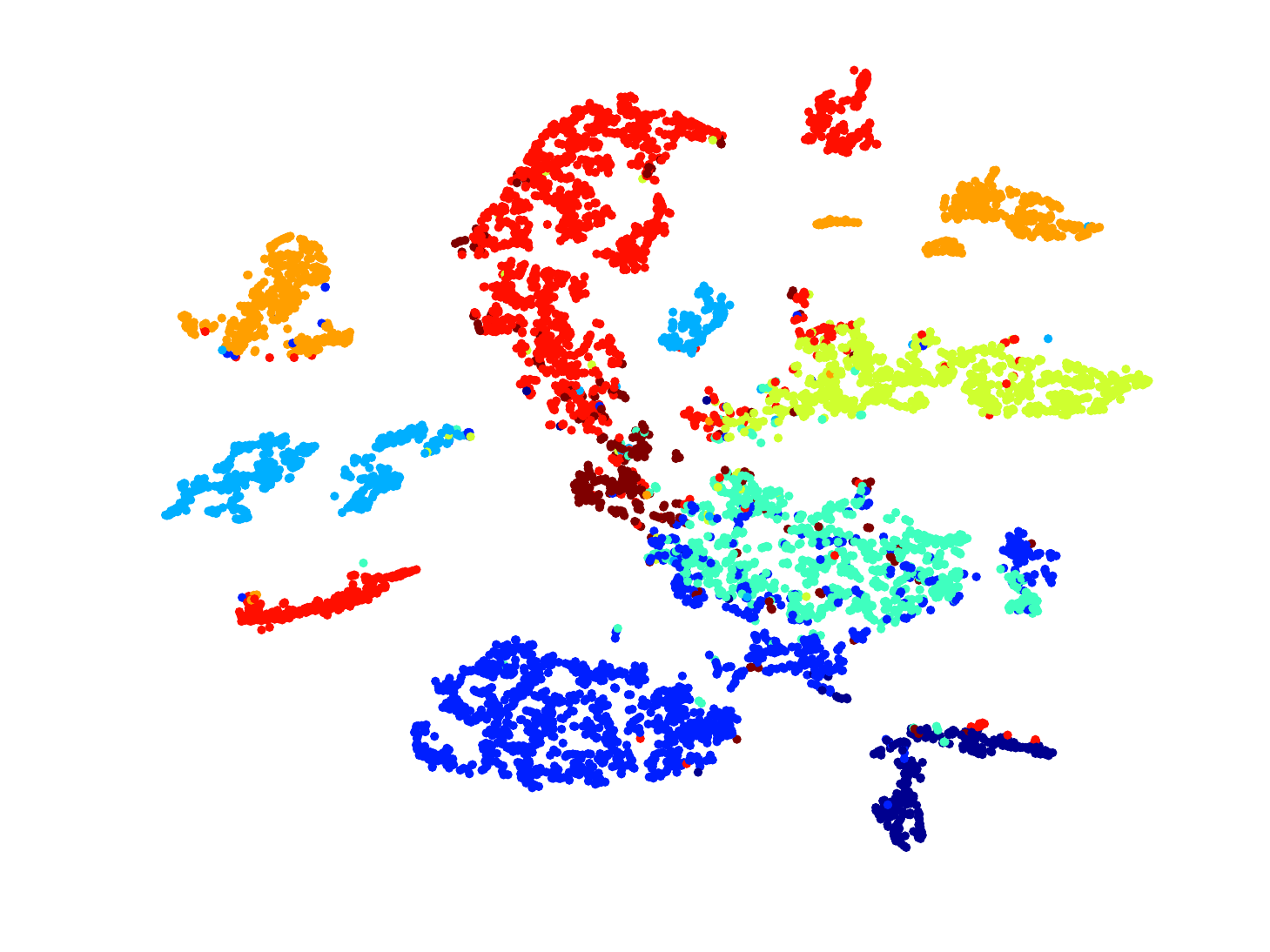}
		\end{minipage}
	}
	\subfigure[GCN-$\mathrm{AGG}_{\ell_p}$]
	{   \begin{minipage}{4.1cm}
			\centering
			\includegraphics[width=1.0\textwidth]{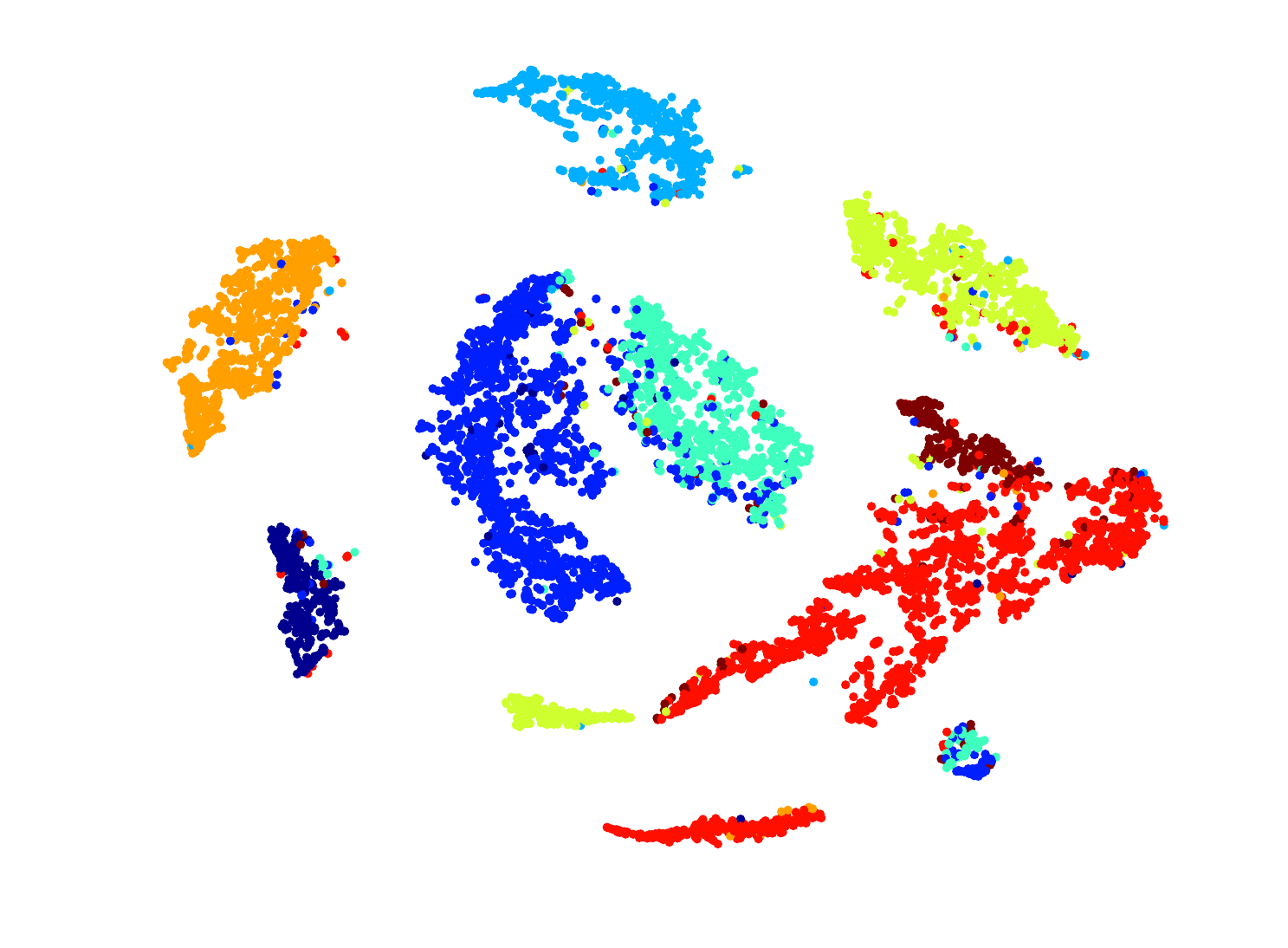}
		\end{minipage}
	}
	\subfigure[GCN-$\mathrm{AGG}_{\mathcal{P}}$]
	{   \begin{minipage}{4.1cm}
			\centering
			\includegraphics[width=1.0\textwidth]{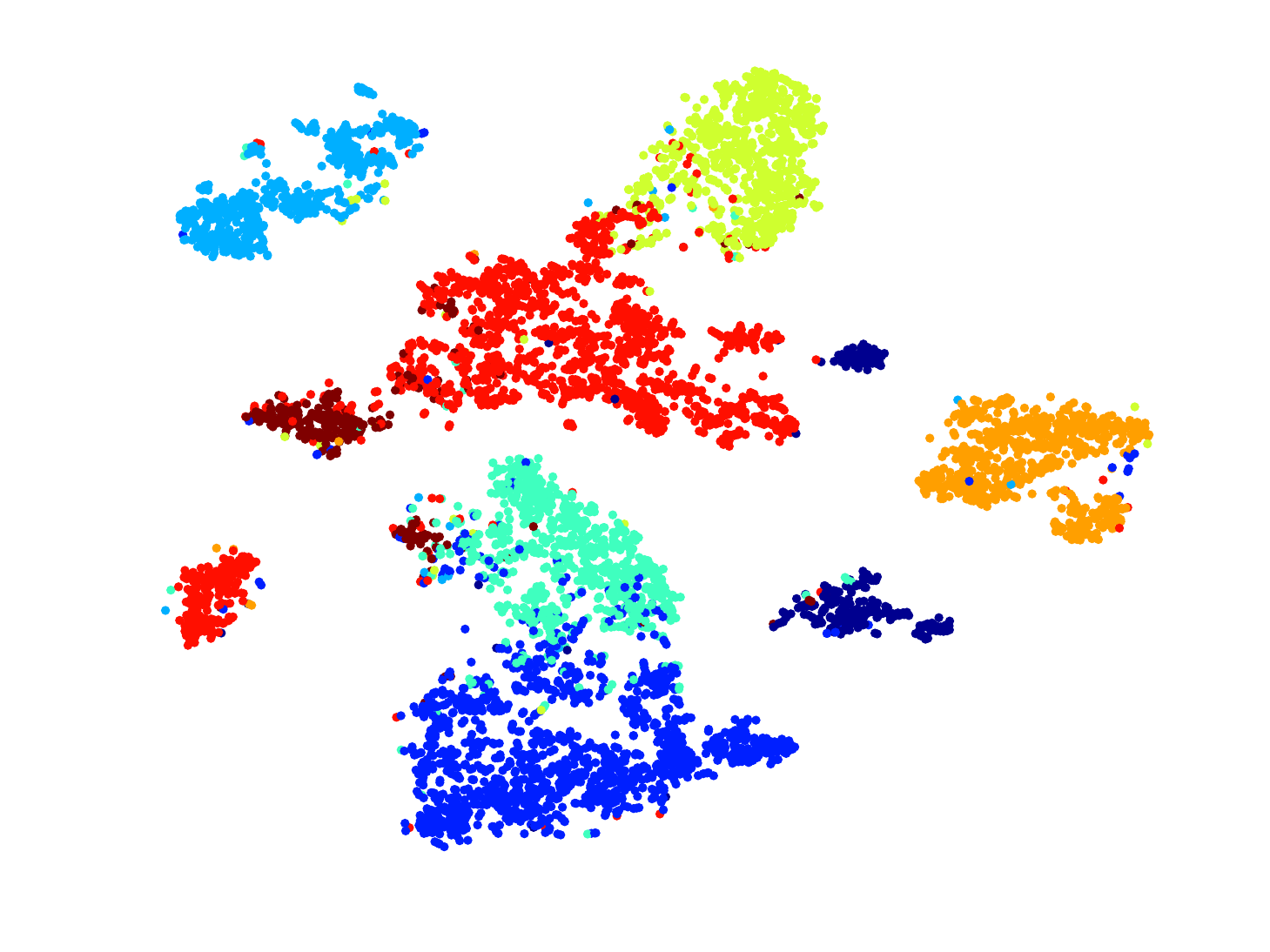}
		\end{minipage}
	}
	\subfigure[GCN-$\mathrm{AGG}_{\mathcal{S}}$]
	{   \begin{minipage}{4.1cm}
			\centering
			\includegraphics[width=1.0\textwidth]{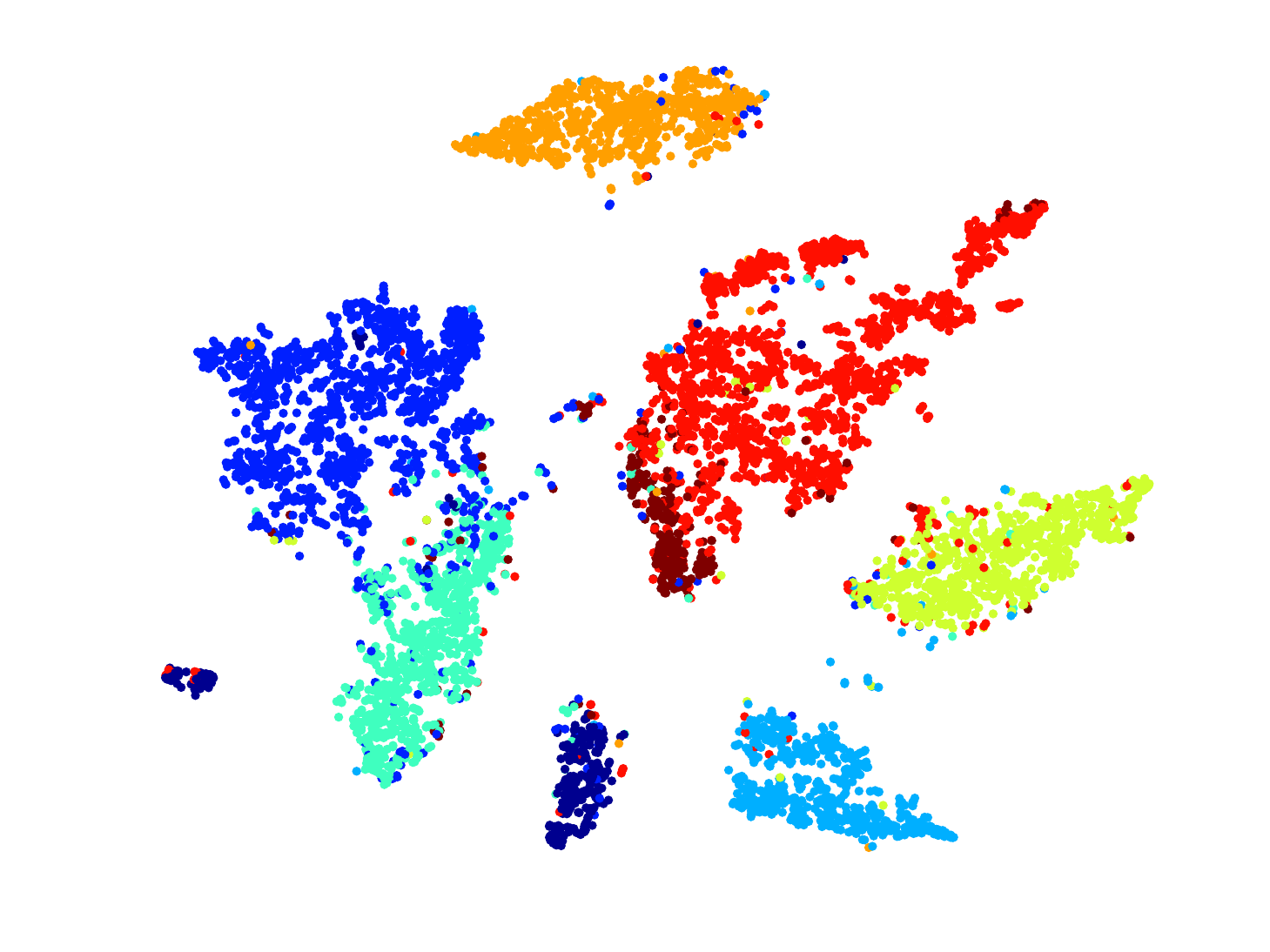}
		\end{minipage}
	}												
	\caption{2D demonstrations of embedding output by GCN~\cite{kipf2017semi} and GCN based on nonlinear aggregations on Photo dataset.}\label{fig::tsne}
\end{figure*}
%%%%
%%%%%
\begin{figure*}[!ht]
	\centering
	\subfigure[GCN]
	{   \begin{minipage}{4.1cm}
			\centering
			\includegraphics[width=1\textwidth]{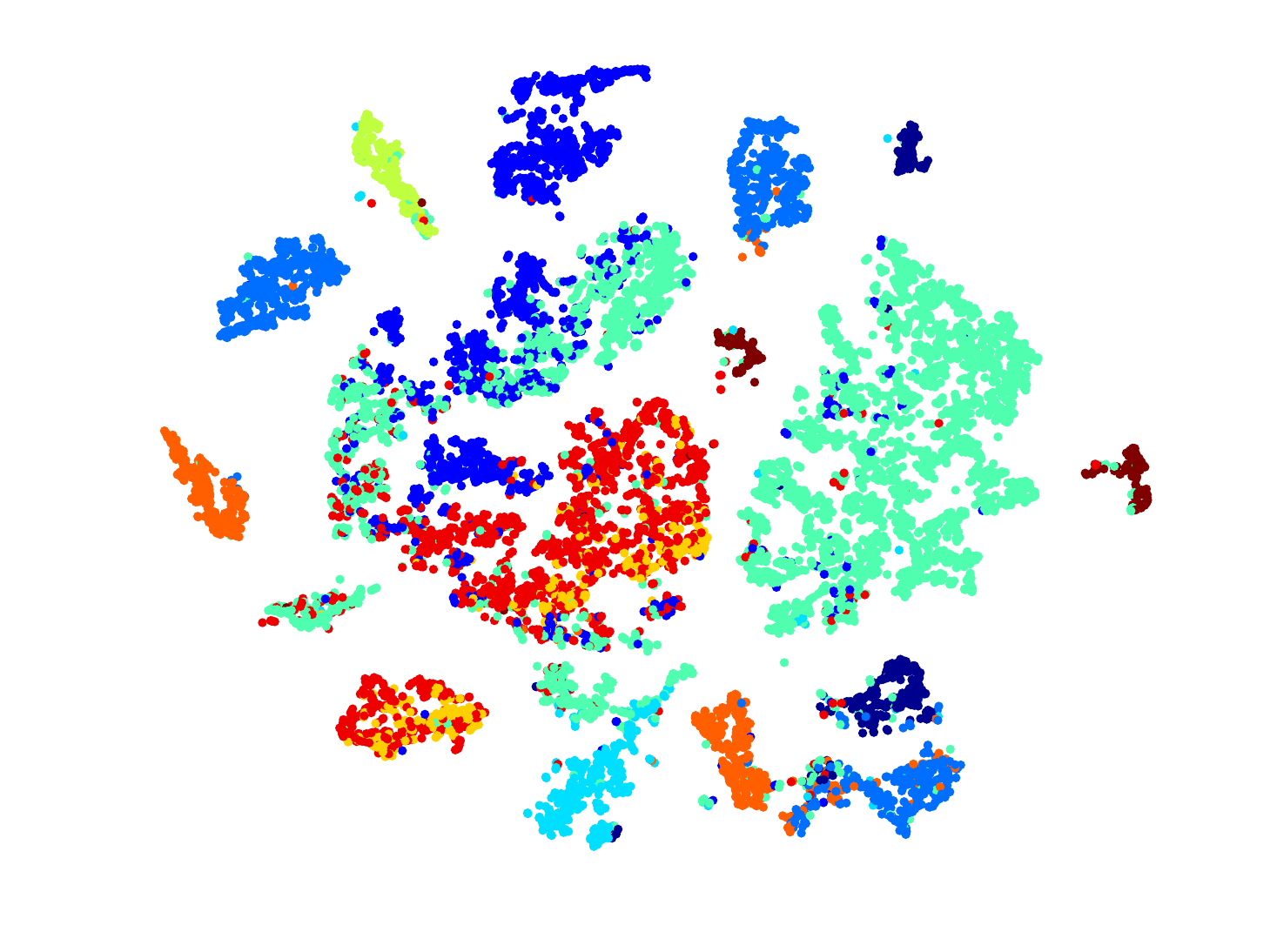}
		\end{minipage}
	}
	\subfigure[GCN-$\mathrm{AGG}_{\ell_p}$]
	{   \begin{minipage}{4.1cm}
			\centering
			\includegraphics[width=1\textwidth]{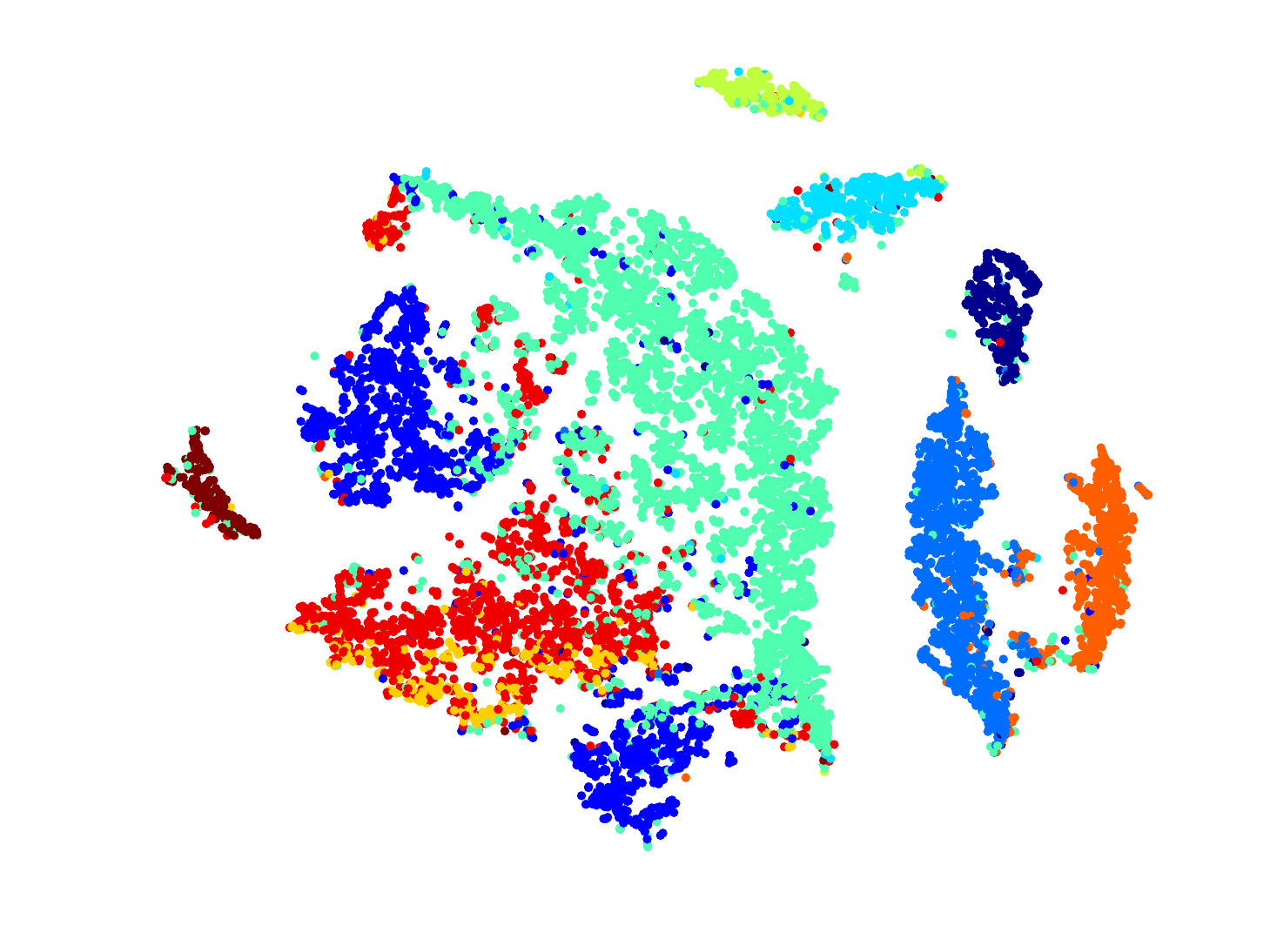}
		\end{minipage}
	}
	\subfigure[GCN-$\mathrm{AGG}_{\mathcal{P}}$]
	{   \begin{minipage}{4.1cm}
			\centering
			\includegraphics[width=1\textwidth]{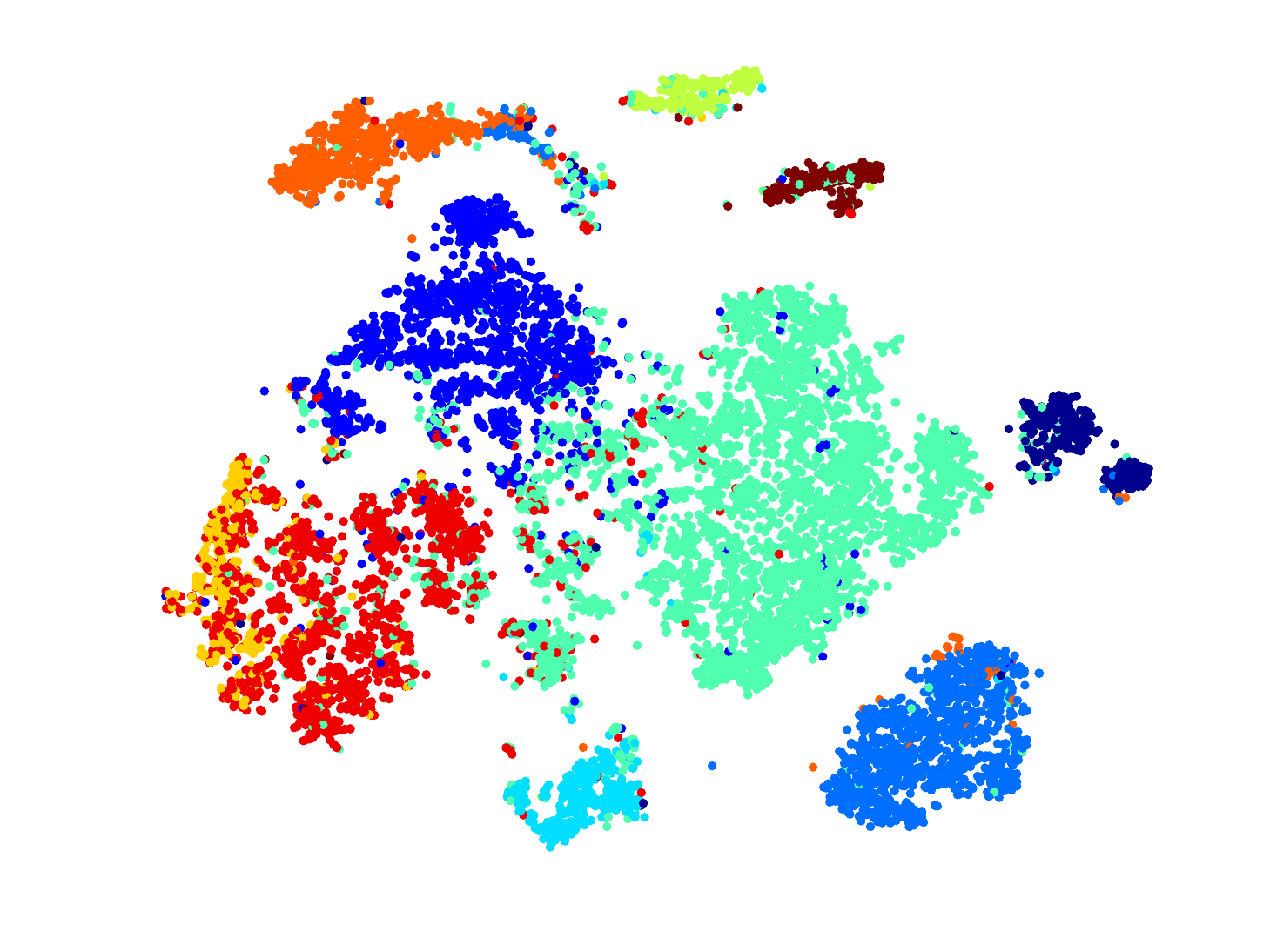}
		\end{minipage}
	}
	\subfigure[GCN-$\mathrm{AGG}_{\mathcal{S}}$]
	{   \begin{minipage}{4.1cm}
			\centering
			\includegraphics[width=1\textwidth]{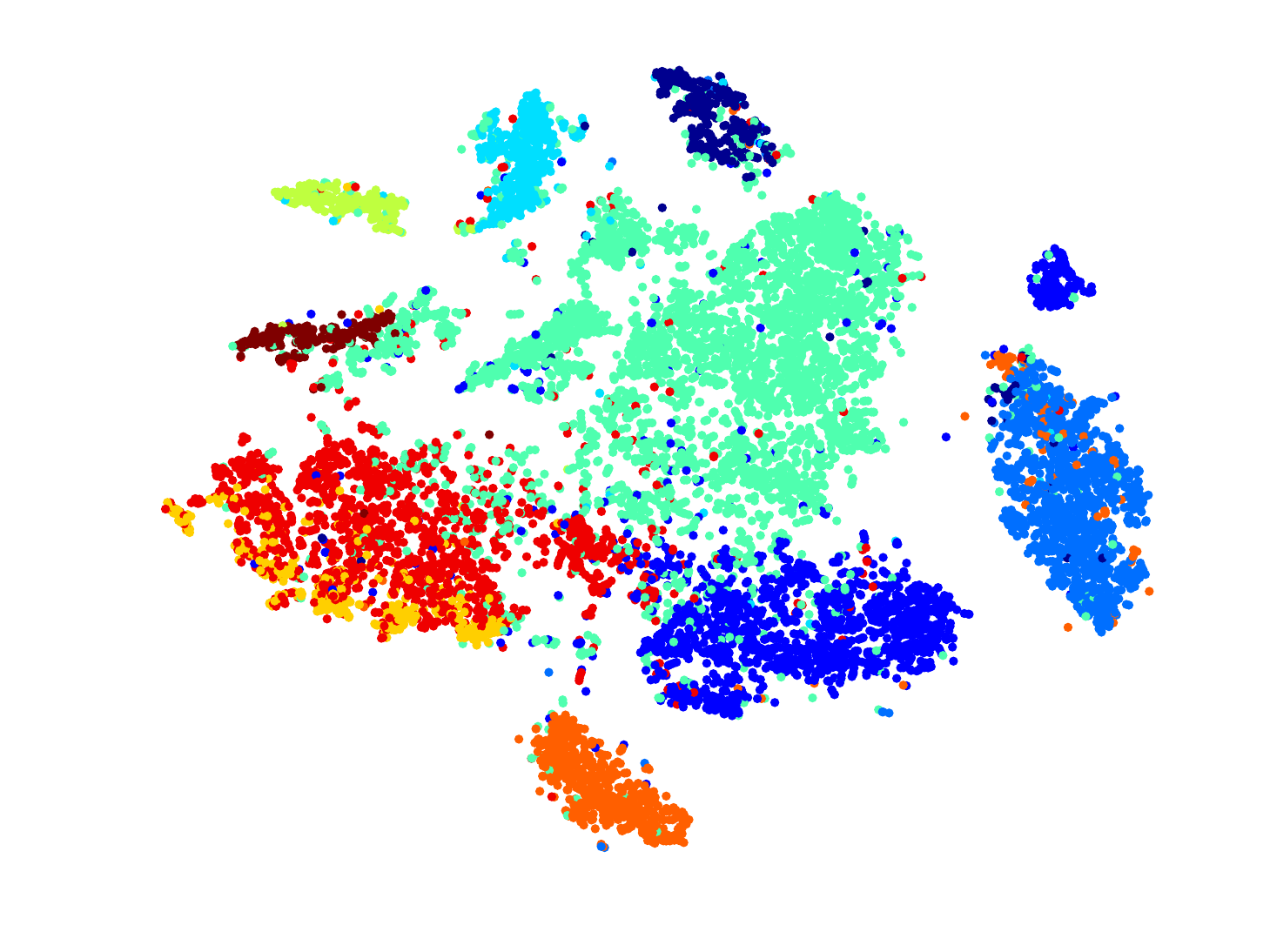}
		\end{minipage}
	}												
	\caption{2D demonstrations of embedding output by GAT~\cite{velickovic2018graph} and GAT based on nonlinear aggregations on Computers dataset.}\label{fig::tsne1}
\end{figure*}
%%%%%
%%------------------------------------
\begin{figure*}[!ht]
	\centering
	\subfigure[Cora]
	{   \begin{minipage}{5.5cm}
			\centering
			\includegraphics[width=1.\textwidth]{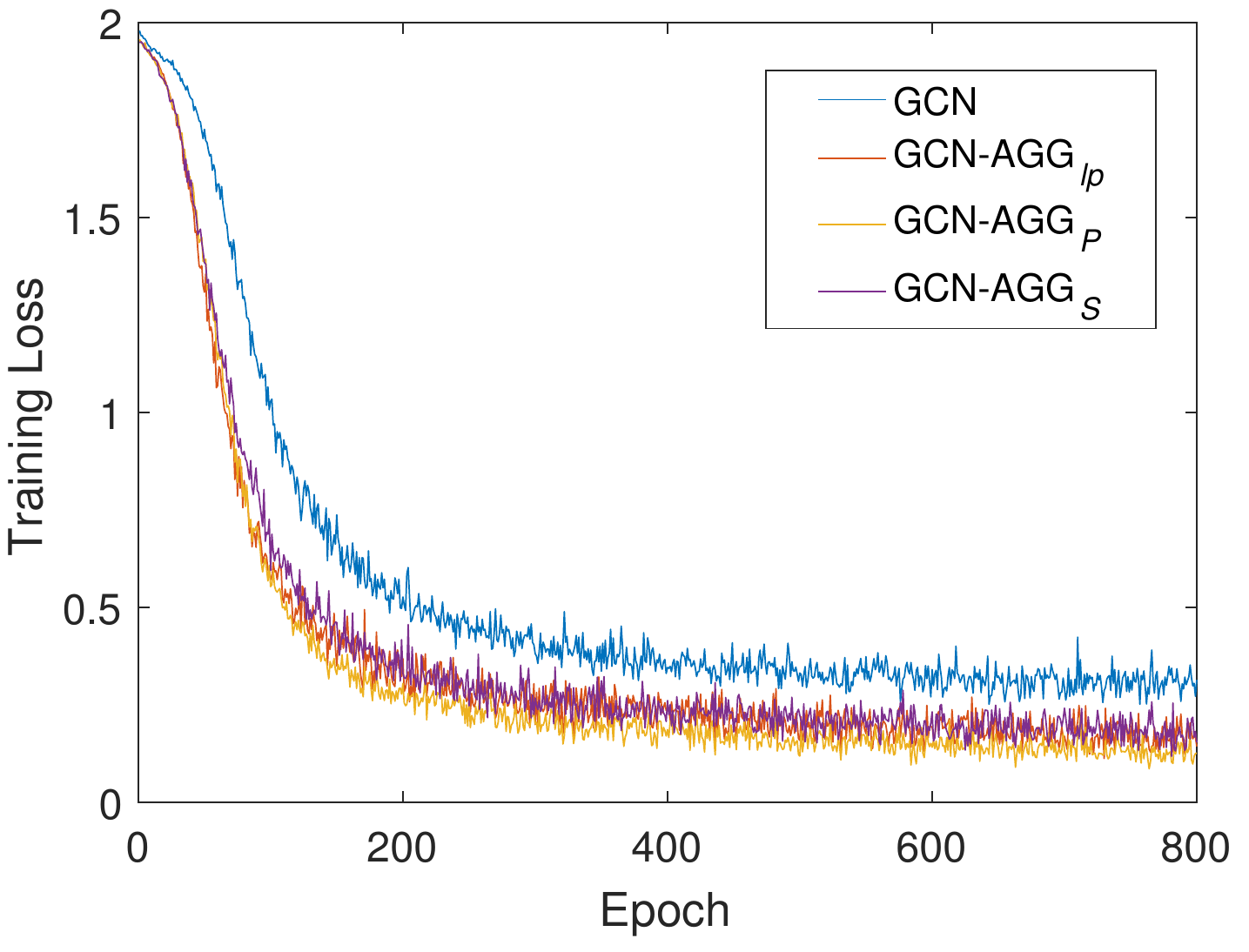}
		\end{minipage}
	}
	\subfigure[Citeseer]
	{   \begin{minipage}{5.5cm}
			\centering
			\includegraphics[width=1.\textwidth]{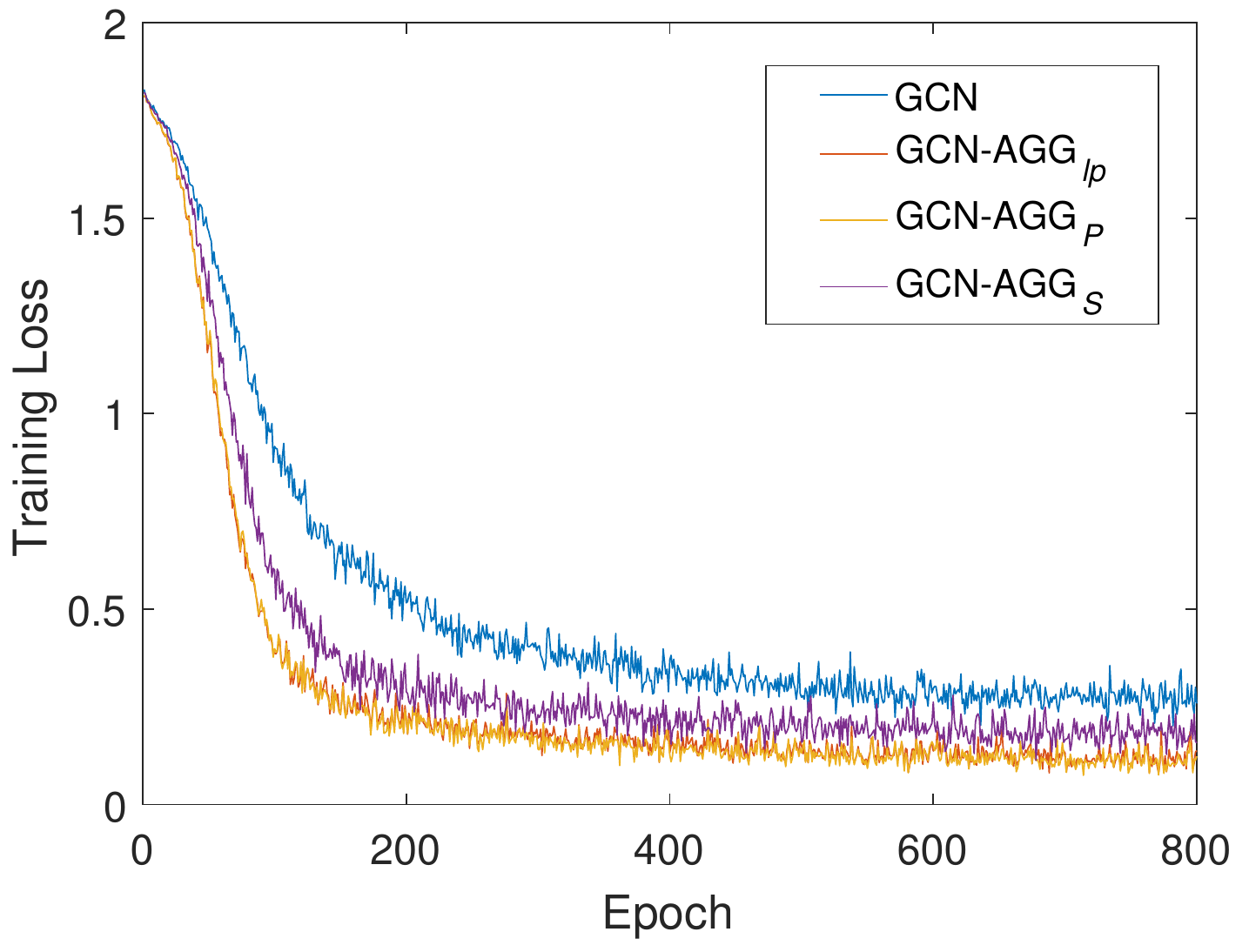}
		\end{minipage}
	}
	\subfigure[Pubmed]
	{   \begin{minipage}{5.5cm}
			\centering
			\includegraphics[width=1.\textwidth]{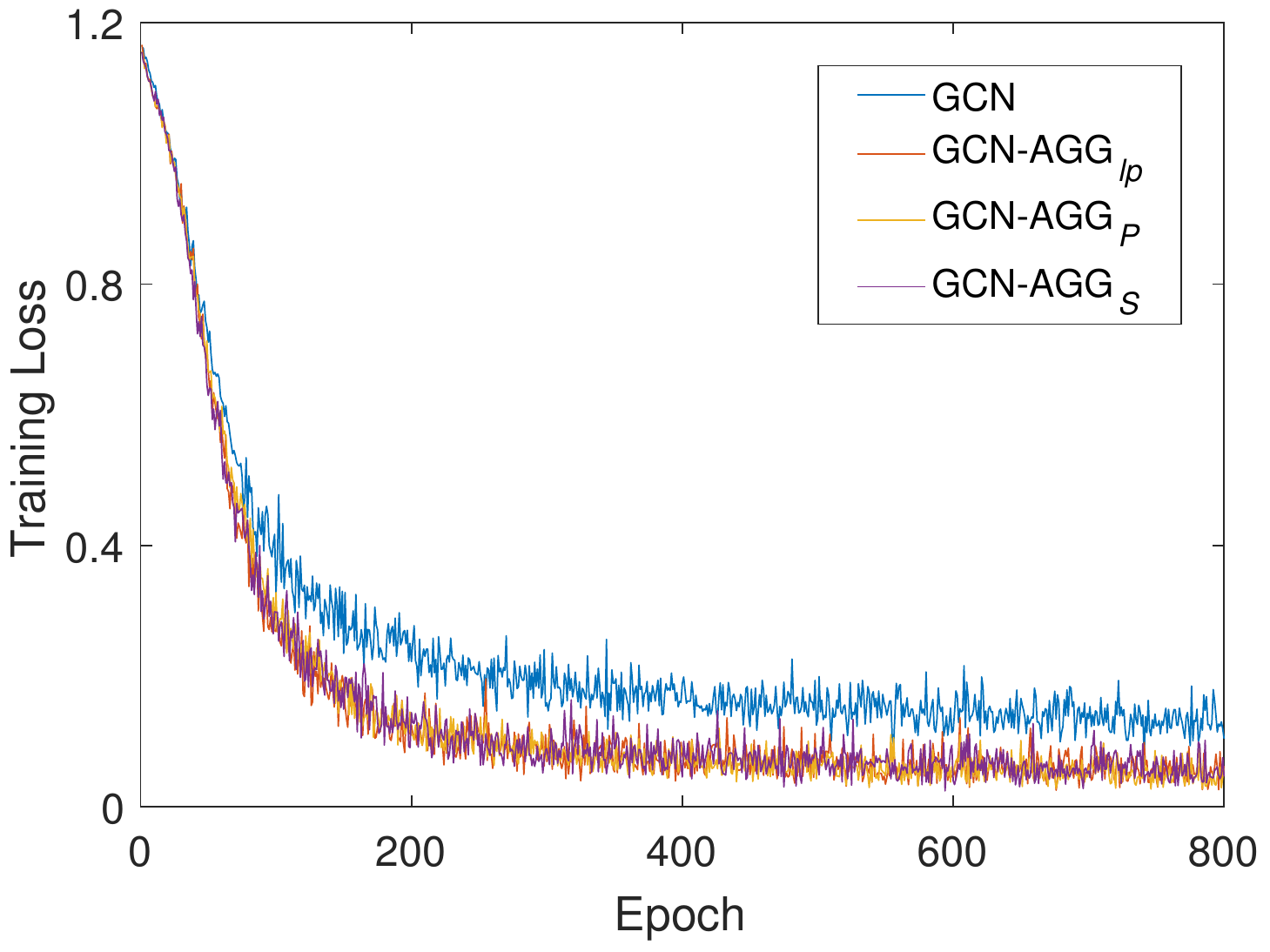}
		\end{minipage}
	}												
	\caption{Comparison results of convergence of GCN~\cite{kipf2017semi} and GCN based on nonlinear aggregations on citation dataset.}\label{fig::loss}
\end{figure*}
%%---------------------------------
%%%%%
%%%%

%\subsection{Parameter Analysis}
%It is known that most deeper GNNs lead to the problem of over-smoothing~\cite{kipf2017semi,li2018deeper,PairNorm}. In this section, we take the nonlinear aggregation based on GCNs without hidden activation as example and show the results of our nonlinear models with different hidden layers in Figure \ref{fig::depth}.
%We can observe that the proposed nonlinear models obtain better results across different hidden layers and consistently outperform the vanilla GCN~\cite{kipf2017semi}.

\subsection{Intuitive Demonstration}

Here, we show some visual demonstrations to intuitively demonstrate the effect of the proposed nonlinear aggregations. We first use 2D t-SNE visualization~\cite{maaten2008visualizing} to show some demonstrations. Figure \ref{fig::tsne} and \ref{fig::tsne1} respectively show the comparison results of GCN~\cite{kipf2017semi} and GCN-based variants  on Amazon Photo and Computers datasets.
As shown in Figure \ref{fig::tsne} and \ref{fig::tsne1}, the node embeddings obtained by our proposed nonlinear GCNs are distributed more compactly and clearly than the vanilla GCN~\cite{kipf2017semi}. It is consistent with the experiment results shown in Table $2$ and $3$ which demonstrates the more expressive capacity of the proposed nonlinear aggregators.
We then show the convergence property of the nonlinear aggregations based on GNNs.
Figure \ref{fig::loss} shows the comparison training loss across different epochs of GCN~\cite{kipf2017semi}, GCN-$\mathrm{AGG}_{\ell_p}$, GCN-$\mathrm{AGG}_{\mathcal{P}}$ and GCN-$\mathrm{AGG}_{\mathcal{S}}$ on citation datasets .
% From Figure \ref{fig::loss},
One can note that our proposed three nonlinear GCNs have lower training loss  which indicates the higher capacity of the proposed aggregators.

\section{Conclusion}

In this paper, we re-think the neighborhood aggregation mechanism of GNNs and propose nonlinear message aggregation schemes to extend GNNs' learning capacity. The proposed nonlinear aggregation operators are general and flexible strategies GNNs which provide the intermediates between the commonly used max and mean/sum aggregations. % which provide optimally balanced aggregation strategies for GNNs.
%
%Thus, our aggregators can inherit both (i) high nonlinearity that increases network's capacity and (ii) detail-sensitivity that preserves the detailed information of representations together in GNNs' message propagation.
%
We integrate our proposed nonlinear aggregators into several GNNs. Experiments on several benchmark datasets show the effectiveness of our proposed nonlinear aggregations on enhancing the learning capacity and performances of GNNs.
%\section{Acknowledgments}

\bibliographystyle{ACM-Reference-Format}
\bibliography{NNA-GNNs}

\appendix

%Lorem ipsum dolor sit amet, consectetur adipiscing elit. Morbi
%malesuada, quam in pulvinar varius, metus nunc fermentum urna, id
%sollicitudin purus odio sit amet enim. Aliquam ullamcorper eu ipsum
%vel mollis. Curabitur quis dictum nisl. Phasellus vel semper risus, et
%lacinia dolor. Integer ultricies commodo sem nec semper.
%Etiam commodo feugiat nisl pulvinar pellentesque. Etiam auctor sodales
%ligula, non varius nibh pulvinar semper. Suspendisse nec lectus non
%ipsum convallis congue hendrerit vitae sapien. Donec at laoreet
%eros. Vivamus non purus placerat, scelerisque diam eu, cursus
%ante. Etiam aliquam tortor auctor efficitur mattis.
%Nam id fermentum dui. Suspendisse sagittis tortor a nulla mollis, in
%pulvinar ex pretium. Sed interdum orci quis metus euismod, et sagittis
%enim maximus. Vestibulum gravida massa ut felis suscipit
%congue. Quisque mattis elit a risus ultrices commodo venenatis eget
%dui. Etiam sagittis eleifend elementum.
%
%Nam interdum magna at lectus dignissim, ac dignissim lorem
%rhoncus. Maecenas eu arcu ac neque placerat aliquam. Nunc pulvinar
%massa et mattis lacinia.

\end{document}